\documentclass[letterpaper, 10 pt, conference]{ieeeconf}  % Comment this line out if you need a4paper

\IEEEoverridecommandlockouts                              % This command is only needed if 
\usepackage[utf8]{inputenc} % allow utf-8 input
\usepackage[T1]{fontenc}    % use 8-bit T1 fonts
\usepackage{hyperref}       % hyperlinks
\usepackage{url}            % simple URL typesetting
\usepackage{booktabs}       % professional-quality tables
\usepackage{amsfonts}       % blackboard math symbols
\usepackage{nicefrac}       % compact symbols for 1/2, etc.
\usepackage{microtype}      % microtypography
\usepackage{xcolor}         % colors

\usepackage{multirow}
\usepackage{wrapfig}
\usepackage{booktabs}
\usepackage{tabularx}
\usepackage{amsmath}
\usepackage{graphicx}

\title{\LARGE \bf
Benchmarking Autonomous Driving Planners Across Leaderboards: A Unified CARLA-Based Evaluation
}

\author{%
 Merve Atasever \quad Alfredo Reina Corona \quad    Zhuochen Liu \quad Qingpei Li   \\ Akshay Hitendra Shah \quad
 Hans Walker \quad Jyotirmoy V.~Deshmukh \quad Rahul Jain \\
\textit{University of Southern California, Los Angeles, California, USA}
 }

\begin{document}

\maketitle
\thispagestyle{empty}
\pagestyle{empty}

%%%%%%%%%%%%%%%%%%%%%%%%%%%%%%%%%%%%%%%%%%%%%%%%%%%%%%%%%%%%%%%%%%%%%%%%%%%%%%%%
\begin{abstract}

 Autonomous driving remains a highly active research domain that seeks to enable vehicles to perceive dynamic environments, predict the future trajectories of traffic agents such as vehicles, pedestrians, and cyclists and plan safe and efficient future motions. To advance the field, several competitive platforms and 
benchmarks have been established to provide standardized datasets and evaluation protocols. 
Each offers a unique dataset and  challenging planning problems spanning a wide range of driving scenarios and conditions. In this study, we present
a comparative case study of representative motion planning methods
drawn from major benchmark ecosystems, including CARLA, nuPlan, and the Waymo
Open Dataset. To ensure a fair and unified evaluation, we adopt CARLA Leaderboard
v2.1 as our common evaluation platform and evaluate eight representative methods:
TF++, InterFuser, TCP, PDM-Lite, MTR+MPC, CaRL, PlanT 2.0, Diffusion planner. By highlighting
the strengths and weaknesses of current approaches, we identify prevailing trends,
common challenges, and potential directions for advancing motion-planning research.

\end{abstract}

%%%%%%%%%%%%%%%%%%%%%%%%%%%%%%%%%%%%%%%%%%%%%%%%%%%%%%%%%%%%%%%%%%%%%%%%%%%%%%%%

\section{Introduction}

Since the DARPA Urban Challenge of 2007, we have been promised that self-driving vehicles are just around the corner. We have been promised Level 5 autonomous driving (AD) technology for the last decade or so; this raises the question: \textbf{\textit{Are we there yet?}} Despite (limited) deployment of test fleets the short answer is {\em no}; the AV (autonomous vehicles) technology is still at the R\&D stage. 
%Any deployed test vehicles require constant human supervision (even if no driver is visible in the vehicle) to avoid various edge cases, and to avoid safety violations. 
Furthermore, almost all current deployments depend on high-definition (HD) maps which require constant remapping, which limits deployment areas of the test fleets. 

To be fair, we have come quite a long way: perception systems aboard AVs have become quite sophisticated and fairly reliable albeit heavily dependent on overly instrumented hardware sensor suite. The planning and control modules are no longer based on rule-based expert systems which had the obvious problem of not being robust to edge/corner cases. And yet, both state estimation and planning and control modules are still at the R\&D stage and not ready for mass deployment. Thus, AV technology development remains an active research field. 

This has spawned three particular research areas: (1) AV simulation which has emphasized extensive sensor simulation, near real-world physics simulation and hyper-photorealism, (2) (open loop) motion prediction of traffic agents, and (3) motion planning for an ego agent while accounting for actions taken by other agents including pedestrians \cite{montali2024waymo, hu2023planning, booher2024cimrl, jia2025drivetransformer, li2024think2drive}. There are a number of AI/ML-based models for motion planning that are regarded as being State-of-the-art (SOTA) models and near the top of various leaderboards. However, their systematic comparison is lacking as different models use different datasets, or different platforms for evaluation. Bench2Drive addresses a related benchmarking problem by evaluating closed-loop end-to-end driving using a curated set of driving abilities \cite{jia2024bench2drive}. Our objective is complementary. Rather than introducing a new benchmark or restricting the comparison to a single family of end-to-end approaches, we study how representative planners originating from different benchmark ecosystems behave when evaluated under the same CARLA closed-loop execution and scoring framework. In particular, our study spans sensor-based end-to-end models, privileged-state planners, rule-based planning, reinforcement learning, imitation learning, prediction $+$ control pipelines, and diffusion-based planning. We additionally evaluate performance at both the map level and the individual traffic-scenario level. This allows us to investigate which sensing, planning, output-representation, and training choices are associated with different types of success and failure under a common evaluation environment.

Many models utilize modular or end-to-end pipelines to deal with  perception, prediction and planning tasks. Those  that implement the modular approach try to solve each task separately. Generally, the pipeline starts with multi-sensor data fusion (e.g., LiDAR, camera, radar, and trajectory data) to perceive the environment, and then continues with transformer-based modules to predict, or plan the ego vehicle's future motion (and the other surrounding agents' motions in some models). Each module uses the output of the previous module in a sequential manner \cite{zhang2023trafficbots, kochanthara2024safety}. However, this leads to compounding errors in which the error of each module accumulates. 
On the other hand, end-to-end AD systems, leveraging mostly reinforcement or imitation learning, have attracted attention  in recent years, where the models aim at optimizing the three tasks together in a unified manner  \cite{chen2024end, chitta2022transfuser, yang2024visual, zhang2021end, buhler2020driving}. Here, the models are trained to minimize a unified loss function, and output decisions for the ego vehicle (or multiple vehicles) directly  \cite{wang2310drive, huang2023differentiable,UniAD}. For both settings, the output can be low level control signals, such as the throttle, speed, steering angle, etc., or can be high-level signals such as planned trajectory, i.e. future waypoints \cite{huang2023applications}. The high-level outputs can then be converted into low-level actuation signals via PID controllers \cite{aastrom2006advanced}, or similar tools. 

Finally, some recent AD research has harnessed multi-modal foundation models and encompassed both large language models (LLMs) and vision-language models (VLMs)
to enhance world modeling, reasoning, and planning \cite{sima2024drivelm, zhang2023trafficbots, hu2023gaia, wang2023drivedreamer}. Although, multi-modal FM approaches promise to improve adaptability and explainability of future AD systems, the full assessment of these models (covering inference latency, prompt engineering, and multi-modal fusion) would significantly expand scope and shift our focus on benchmarking core planning algorithms.

Waymo, nuPlan, and CARLA leaderboards serve as prominent platforms that advance autonomous-driving (AD) research by providing standardized benchmarks which ensure accelerating progress in simulation, motion prediction, and motion planning. However, it is hard to cross-compare top performers on Waymo and nuPlan leaderboards leading to a doubt whether the models have been optimized to hill-climb for those specific benchmarks. Thus, in this work, we perform a detailed comparative analysis of SOTA models drawn from these leaderboards, as well as CaRL, PlanT 2.0, and Diffusion planner on the CARLA Leaderboard v2.1 evaluation platform \cite{Beißwenger2024PdmLite,jaeger2025carl,gerstenecker2025plant, zheng2025diffusionbased}. %We first compiled all top-ranked methods from each competition since 2022, then excluded those lacking publicly available code and pretrained weights to arrive at a final shortlist of eight methods. 
On the Waymo leaderboard, most leading methods adopt the Motion Transformer (MTR) architecture as their backbone, elaborating on it with task-specific enhancements \cite{shi2022mtr, wang2023multiverse}. Because the original MTR only predicts the future trajectories of the ego vehicle and its seven nearest neighbor agents, we augment it with an MPC-based planning module. Thus, performance of MTR+MPC model is being reported for the first time. Details of this integrated pipeline are provided in the following section, and the full list of evaluated methods can be found in Table \ref{modeldetails}.

The main contributions of our paper can be summarized as follows:

\noindent $\star$ \textbf{Unified Benchmarking:} We systematically examine eight
representative motion planning methods from CARLA, Waymo Open Dataset,
and nuPlan ecosystems on the CARLA Leaderboard v2.1 evaluation
platform.

\noindent $\star$ \textbf{Insightful Analysis:} We quantitatively evaluate and report
each method's strengths and weaknesses across different traffic scenarios and maps.

\noindent $\star$ \textbf{Roadmap for Future Research:} We uncover common trends and
failure modes of current planners, highlighting concrete research directions for
robust AD motion planning.

\noindent $\star$ \textbf{The MTR+MPC Model:} We augment the Motion Transformer (MTR)
backbone, originally designed for forecasting future trajectories, with an MPC-based
planning module and evaluate this integrated pipeline in closed-loop setting.

\noindent $\star$ \textbf{Compatibility with CARLA 0.9.15:} We refactor the old
codebases of InterFuser, TF++, and TCP to be compatible with CARLA 0.9.15, enabling
training and testing on the newly added maps and scenarios \cite{shao2023safety, wu2022trajectory, jaeger2023hidden}. This contribution enables
the community to reuse these models in the latest CARLA evaluation setting. All of
this will be made available in a public GitHub repository.

\section{Models for Autonomous Driving}

%Motion planning research has branched into two predominant approaches: modular
%pipelines and end-to-end learning systems. Both methodologies incorporate their
%respective architectural advances. The separation of tasks affords interpretability and the ability to mix and match best components, but suffers from compounding errors and decrease performance in some scenarios. End-to-end systems, by contrast, collapse these stages into a single network trained with imitation or reinforcement learning, optimizing a joint objective to map raw sensor inputs to output commands or trajectories. While they can better handle long-tail cases and extract task-specific features, they lack interpretability and can be difficult to tune. Thus, hybrid strategies have emerged as well: transformer-based backbones for multimodal scene understanding and  trajectory prediction with rule-based controllers that inject expert knowledge. This hybrid scheme bridges high-level planning with low-level control. Our study helps to illustrate how these branches work in state-of-the-art methods and remark that we focus on the established baselines that currently define the performance limits on major leaderboards.

\begin{table*}[t]
  \caption{Architectural Details of Benchmark Models.}
  \label{modeldetails}
  \vskip 0.15in
  \centering
  \small
  \begin{tabularx}{\linewidth}{@{} l 
      p{3cm} 
      p{6cm} 
      X            % this X column will flexibly wrap
      @{}}
    \toprule
    \textbf{Model} 
      & \textbf{Input Data} 
      & \textbf{Model Architecture} 
      & \textbf{Planning Representation} \\
    \midrule
 TF++    & RGB camera $+$ LiDAR & Transformer-based multimodal input fusion / Supervised imitation learning & Spatial path $+$ target-speed distribution \\
Interfuser & Multi-view RGB $+$ LiDAR & Transformer encoder-decoder / Supervised learning from expert demonstrations & Waypoints $+$ object density $+$ traffic rule representations \\
TCP   & Front RGB $+$ vehicle measurements & CNN/GRU dual-branch network / Imitation learning $+$ auxiliary objectives &  Waypoints $+$ direct control predictions \\
PDM Lite  & Privileged map and dynamic-actor states & Rule-based planner & Candidate motion and target-speed logic \\
MTR + MPC  & Map $+$ agent history & Motion Transformer $+$ MPC / Prediction learning $+$ online optimization & Predicted multi-agent trajectories used by MPC\\
CaRL  & Privileged BEV semantic state & CNN policy / Reinforcement learning & Direct policy output \\
PlanT 2.0  & Privileged object states $+$ road-layout information & Object-centric transformer / Imitation learning & Spatial path-points $+$ temporal waypoints\\
Diffusion-Planner & Structured map, actor states, and histories &  DiT-based diffusion planner /  Diffusion-based trajectory generation & Ego trajectory jointly modeled with nearby-agent predictions \\
    \bottomrule
  \end{tabularx}
  \vskip -0.1in
\end{table*}

%Table~\ref{modeldetails} provides a compact comparison of each method's observation space, learned architecture, training paradigm, and planning representation.

\noindent\textbf{TransFuser++.} TF++ is a sensor-based end-to-end planner using a front RGB camera and 360° LiDAR. Separate convolutional encoders process the modalities before transformer-decoder attention performs multimodal fusion. Importantly, TF++ separates lateral and longitudinal planning by predicting a deterministic spatial path together with a target-speed distribution, which are subsequently converted to low-level controls using a PID controller.

\noindent\textbf{InterFuser.} InterFuser combines multi-view RGB and LiDAR inputs using a transformer architecture. Its decoder represents multiple driving-relevant quantities, including future waypoints, object-density information, and traffic-rule information. These predictions are passed to an explicit safety module and subsequently to a PID controller, creating a hybrid interface between learned scene reasoning and structured low-level control.

\noindent\textbf{Trajectory-guided Control Prediction.}
TCP is a camera-based end-to-end planner that jointly learns trajectory and low-level control prediction. A GRU-based trajectory branch predicts future waypoints, while a control branch directly predicts driving controls. During inference, the contributions of the two branches are adjusted according to the driving context, providing a useful comparison with methods that rely exclusively on waypoint tracking or exclusively on direct control.

\noindent\textbf{Oracle Baseline (PDM-Lite).} PDM-Lite is a privileged rule-based planner that receives ground-truth simulator state, route information, and dynamic-agent information. It combines rule-based target-speed selection, short-horizon actor rollout, collision checking, and low-level steering and longitudinal control. We use it as a strong privileged rule-based reference rather than as a strict performance upper bound.

\noindent\textbf{CaRL.}
CaRL is a privileged reinforcement-learning planner trained using PPO. Its reward design emphasizes route completion, while infractions are incorporated through episode termination or multiplicative reductions in route-completion reward. CaRL therefore provides a useful comparison between learned reinforcement-learning behavior and privileged rule-based or imitation-learning planning.

\noindent\textbf{PlanT 2.0.} PlanT 2.0 is a privileged object-centric transformer planner that receives ground-truth object states, route information, speed limits, and semantic road-layout information. It predicts spatial path-points for lateral planning and temporal waypoints for longitudinal planning, which are tracked using low-level controllers adapted from PDM-Lite.

\noindent\textbf{Diffusion-Planner.} Diffusion-Planner is a structured trajectory planner originating from the nuPlan ecosystem. It conditions on ego state, neighboring-agent history, vectorized lane information, static objects, and route information, and jointly models the ego trajectory together with nearby-agent predictions using a diffusion-transformer architecture. For CARLA evaluation, we keep the pretrained network fixed and replace its original preprocessing interface with a CARLA-to-nuPlan conversion layer.

\noindent\textbf{MTR+MPC.} Motion Transformer \cite{shi2022motion} (MTR) is an encoder-decoder architecture for predicting multimodal behavior of traffic agents and is ranked \# 1 in Waymo Open Dataset Motion Prediction Challenge in 2022.
Since its introduction, most leading motion prediction methods in the Waymo challenge have adopted MTR as their core backbone. 
Notable variants include MTR-A,  MTR++, and MTR v3 \cite{shi2022mtr,shi2024mtr++,shimtrv3}. Building on these advances, 
we integrate a model-predictive control (MPC) planner on top of the MTR backbone to address motion planning tasks. Our modular design allows any alternate
multi-agent prediction model to replace MTR with minimal adaptation.
We trained the MTR model from scratch, while the pre-trained models are available publicly for all other benchmarked methods. Since MTR is trained on the Waymo Open Motion Dataset (WOMD) \cite{sun2020scalability}, we then extract CARLA simulator data and convert it to the same input format. 
MTR predicts the future trajectories of the ego vehicle and its seven nearest neighbor vehicles over an 80-step (eight-second) horizon. 
We use the first N steps of these predictions (the receding horizon) to define the constraints and cost function for our model-predictive control (MPC) module, which outputs the final low-level control commands of steering angle and acceleration. For state 
$x_k$ and control $u_k$ variables, we solve the optimization given below by leveraging CasADi and its IPOPT nonlinear programming solver:
%\vspace*{-1pt}
\begin{align}
\min_{u_{0:N-1}} J(u_{0:N-1}) = J_{\text{terminal}} + \sum_{k=0}^{N-1} \Big( &J_{\text{waypoint},k} \\ + &J_{\text{speed},k} + J_{\text{steering},k} \nonumber \\ + & J_{\text{obs},k} + J_{\text{long},k} \Big) \nonumber
\end{align}
\noindent subject to the following constraints:
\vspace*{-2pt}
\begin{align}
& \text{Dynamics:} \quad x_{k+1} = f(x_k, u_k),  \\
& \text{Steering bounds:} \quad -\delta_{\max} \leq \delta_k \leq \delta_{\max}, \\
& \text{Acceleration bounds:} \quad a_{\min} \leq a_k \leq a_{\max}, \\
& \text{Speed limit:} \quad v_{k+1} \leq v_{\max} + 0.5, 
\end{align}
\noindent for $ k = 0, \ldots, N-1$ and $ x_0 = x_{\text{init}}$. The objective function combines terms for: (i) the terminal deviation from the destination at the end of the receding horizon, (ii) adherence to reference waypoints (i.e., remaining on the road), 
(iii) velocity and steering‐rate smoothness, and (iv) collision cost with neighboring vehicles and static obstacles.
\cite{abbas2017obstacle, liu2017path, yi2018model}. 
The collision term is evaluated using MTR’s predicted trajectories for surrounding traffic agents. Handling traffic infractions such as red lights, stop signs, and emergency vehicles, can be achieved in various ways. One common approach encodes these rules as MILP constraints for a linear solver. In our implementation, however, we model them as static obstacles and incorporate them directly as we leverage IPOPT nonlinear solver. 

\begin{figure}[h]
\centering
\includegraphics[width=0.6\linewidth]{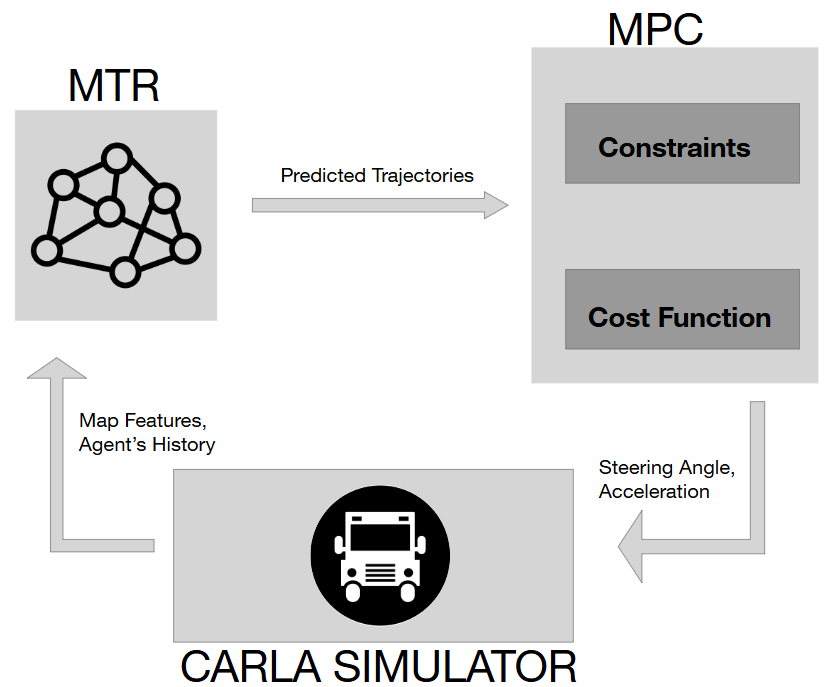}
\caption{MTR+MPC Pipeline.}
 \label{fig:mpc_pipeline}
\end{figure}

\section{Overview of Leaderboards}

\noindent\textbf{The CARLA Leaderboard.}
The Carla Leaderboard \cite{carla_leaderboard_v2_0} is a closed loop metric used to measure agents' performance on routes within Carla. It consists of 2 tracks, SENSORS and MAP track. On the SENSORS track, the agent is given access to the standard AV sensors, such as camera, lidar, radar, etc. On the MAP track, it is additionally given access to an HD map of the area. Agents are evaluated based on a driving score which is the product of two scores: The route completion score, which measures the percent completion of the given route, and the infraction penalty, which is a product of all the infractions the agent commits.

\noindent\textbf{The Waymo Leaderboard.}
The Waymo Leaderboard \cite{ettinger2021large} evaluates motion prediction models on a large-scale dataset comprising over 100,000 scenes, each 20 seconds long at 10 Hz, totaling more than 570 hours of unique data across six cities. Each scenario provides 1 second of agent history and requires models to predict the future positions of up to 8 predicted objects per scenario over an 8-second horizon. Inputs include rich semantic data such as 3D bounding boxes, map features (lanes, crosswalks, stop signs), and agent metadata (type, velocity, heading). Evaluation metrics include Minimum Average Displacement Error (minADE), Minimum Final Displacement Error (minFDE), and Miss Rate assessing the accuracy and feasibility of the predicted trajectories. \textit{Remark:} While Waymax provides dedicated planning baselines, it does not provide an open-source evaluation server. We thus leverage WOMD’s MTR whose code is public and extend it with MPC to study the motion planning \cite{gulino2023waymax}.

\noindent\textbf{The nuPlan Leaderboard.}
The nuPlan Leaderboard \cite{caesar2021nuplan} is a closed-loop planning benchmark that evaluates motion planners on 15-second scenario roll-outs sampled from $\approx$300-hours of real-world data spanning four cities. It supports three tracks: open-loop, closed-loop non-reactive, and closed-loop reactive. Each supplies a 10 Hz top-down semantic state (two seconds of history, routed goal, nearby agents, and HD-map layers) from which the planner must generate a trajectory. Post-simulation, the agent is scored on four things: safety (collisions, time-to-collision), rule compliance (drivable area, direction, speed), efficiency \& progress, and comfort (jerk, yaw-rate, etc.). Safety serves as a hard mask; the weighted average of the remaining metrics yields a per-scenario score in [0, 1], and the leaderboard rank is the mean of these scores across all hidden-test scenarios.

\section{Experimental Results}

The CARLA Autonomous Driving Leaderboard provides an open, reproducible benchmark for assessing the driving performance of autonomous agents in realistic traffic scenarios. In Leaderboard v2.1, each agent is tasked with navigating predefined routes, specified via GPS coordinates and map waypoints, while encountering several traffic scenarios drawn from the NHTSA pre-crash typology\footnote{National Highway Traffic Safety Administration: "Pre-crash scenario typology for crash avoidance research."}. 
 This standardized evaluation framework enables fair comparison across diverse approaches by initializing agents at a common start point and measuring their ability to reach specified destinations under specific conditions.

\subsection{Benchmarks}
For a rigorous assessment of model performance, we executed each model on a diverse suite of benchmark datasets, comprising the CARLA Leaderboard 2.1 scenarios, the CARLA 42 Routes benchmark, the Town05, and the longest6 benchmarks.

$\star$ \textbf{Leaderboard 2.1 scenarios:} The official CARLA Leaderboard dataset provides multiple instances of 10 common traffic scenarios selected from the NHTSA pre-crash typology. Our model evaluation is conducted using the scenario set distributed in the dev-test partition.

$\star$ \textbf{CARLA 42 routes:} The CARLA 42 routes benchmark is proposed in \cite{chitta2021neat}. The routes are collected from Town01-06 with unique combinations of weather and daylight conditions. Additionally, the multi-lane road layouts, distant traffic lights, and high density of background agents make navigation more challenging.

$\star$ \textbf{Town05:} The Town05 benchmark is introduced in \cite{prakash2021multi}. It is a complex town with multi-lane roads, single-lane roads, bridges, highways and exits. The dataset has two settings: (1) Town05 Short consisting of 10 short routes of 100-500m comprising 3 intersections each, and (2) Town05 Long consisting of 10 long routes of 1000-2000m comprising 10 intersections each. Each route consists of a high density of dynamic agents and adversarial scenarios which are spawned at predefined positions along the route.

 $\star$ \textbf{Longest6:} The longest6 benchmark is introduced in \cite{chitta2022transfuser} to balance the map coverage in the official dataset. The benchmark preserves the six longest trajectories drawn from Town01–06 and supplements them with novel route-design choices. Each trajectory spans roughly 1.5 km, closely aligning with the leaderboard’s average length of $\sim$1.7 km, while featuring elevated traffic density and more demanding pre-crash traffic scenarios.

\subsection{Map Data Preprocess for MTR}
All benchmark datasets referenced in this work are derived from the CARLA simulator. By contrast, the MTR model is trained on the Waymo Open Dataset, whose map representation diverges fundamentally from CARLA. Specifically, CARLA encodes road geometry in the ASAM OpenDRIVE specification, whereas the Waymo Open Dataset provides a vectorized map expressed in the Lanelet2 \cite{poggenhans2018lanelet2} standard. To make MTR compatible with CARLA, we convert every CARLA town to Lanelet2: the original OpenDRIVE files are parsed to extract the road network, after which we reconstruct all road and lane topologies as well as associated traffic-control elements in the Lanelet2 schema.

\subsection{Evaluation}

We selected the CARLA Leaderboard v2.1 as our unified evaluation platform because its default metrics provide a balanced assessment of both route completion and infraction penalties. Additionally, the CARLA Leaderboard and its variants are the only established benchmarks for sensorimotor planning (by contrast, Waymax and nuPlan target privileged-information motion planning). While other simulators exist, none offers comparable flexibility, infrastructure, or breadth of evaluated methods. In the CARLA Leaderboard, agent performance is evaluated using three complementary metrics, each computed per route and then averaged to produce an overall score. Because routes may encompass multiple traffic scenarios, we first decompose each route into its individual scenario segments, scrape and evaluate each scenario separately, and then aggregate both route-level and scenario-level results. This two-tiered analysis provides more precise root-cause insights into which models excel or underperform in specific driving conditions. Routes taken from the Leaderboard 1.0 were converted into the 2.1 format directly.

$\star$ \textbf{Driving Score:} \( 
      \text{DS}_i = R_i \, P_i \)
    where \(R_i\) is the percentage of completion of and \(P_i\) is the infraction penalty of the $i^{th}$ route. The driving score is the main metric of the leaderboard and ranges from \( 0 \) to \( 100 \). 

$\star$ \textbf{Route Completion:} 
    \(
      \text{RC}_i = 100 \times \frac{\text{distance traversed}}{\text{total route distance}},
    \)
    i.e. the percentage of the planned route successfully covered.

$\star$ \textbf{Infraction Penalty:} 
    \(
      \text{IP}_i = \frac{1}{1 + \sum_j c_j \cdot n_{ij}}
    \)
    where \(n_{ij}\) is the number of infractions of type \(j\) committed on route \(i\). 
    
For all of the above metrics, higher values correspond to better performance. All benchmarks are run on the same machine with a 32-thread Intel Xeon Gold 6130 CPU, an NVIDIA Quadro P5000 GPU, and 128 GB physical memory.

\subsection{Results}

Table \ref{tab:aggregated-results} summarizes aggregate performance at two complementary levels. The scenario-level results average performance across the individual Leaderboard traffic scenarios, whereas the map-level results average performance across 42Routes, Town05 Long, Town05 Short, and Longest6. We use the two views jointly because they probe different aspects of closed-loop behavior: scenario-level evaluation emphasizes local responses to specific interactions, while long map-level routes additionally expose error accumulation, sustained route following, and repeated interaction over longer rollouts. We therefore analyze not only the final Driving Score (DS), but also its Route Completion (RC) and Infraction Penalty (IP) components to determine whether poor performance is caused primarily by insufficient progress or by rule and safety violations.

\begin{table}[t]
\centering
\caption{Aggregated results. DS: Driving Score, RC: Route Completion, IP: Infraction Penalty.}
\label{tab:aggregated-results}

\scriptsize
\setlength{\tabcolsep}{2.5pt} % default is 6pt
\renewcommand{\arraystretch}{1.0}

\resizebox{\columnwidth}{!}{%
\begin{tabular}{lcccccc}
\toprule
Model & \multicolumn{3}{c|}{Scenario} & \multicolumn{3}{|c}{Map} \\
 & DS \( \uparrow \) & RC \( \uparrow \) & IP \( \uparrow \) & DS \( \uparrow \) & RC \( \uparrow \) & IP \( \uparrow \) \\
\midrule
TF++       & \textbf{55.9} & 70.1 & \textbf{0.76} & 53.6 & 70.5 & 0.69 \\
InterFuser & 25.5 & 42.2 & 0.45 & 25.4 & 54.3 & 0.45 \\
MTR+MPC    & 30.4 & 68.5 & 0.46 & 20.4 & 89.8 & 0.20 \\
PDM-Lite   & 53.1 & \textbf{76.0} & 0.66 & 72.0 & 96.4 & 0.76 \\
TCP        & 14.7 & 34.5 & 0.46 & 21.9 & 48.4 & 0.44 \\
CaRL       & 42.8 & 67.6 & 0.69 & 74.4 & 84.7 & \textbf{0.87} \\
PlanT 2.0     & 53.0 & \textbf{75.7} & 0.61 & \textbf{75.1} & \textbf{97.1} & 0.77 \\ 
Diffusion-Planner & 29.8 & 54.9 & 0.53 & 27.6 & 52.9 & 0.55 \\
\bottomrule
\end{tabular}%
}
\vspace{-2mm}
\end{table}

In the CARLA leaderboard, traffic scenarios are organized into the following five subcategories:

\textbf{Control:} In these scenarios the ego-vehicle loses control due to adverse road or encountering an obstacle, and must recover to its original lane. 

\textbf{Parking:} Here, the vehicle must exit a parallel parking bay in the flow of traffic. 

\textbf{Traffic Negotiation:} These are combinations of left or right turns at signalized or nonsignalized intersections, where the opposite Vehicle may have priority, or may have run the red light.

\textbf{Obstacle Avoidance:} These scenarios encode different kinds of obstacles in the ego-vehicle's lane such as: construction, parked vehicles or their doors, slow-moving hazards, etc. 

\textbf{Braking and Lane Changing:} This includes scenarios for dynamic object crossing, vehicle turning routes which may encounter pedestrians.

\begin{table*}[t]
\centering
%\scriptsize
\setlength{\tabcolsep}{2.2pt}
\renewcommand{\arraystretch}{0.80}
\caption{Detailed driving-scores (DS) for each map and traffic scenario.}
\label{tab:scenario-driving-scores}
%\resizebox{\textwidth}{!}{%
\begin{tabular}{lrrrrrrrr}
\toprule
Traffic scenarios or maps & PDM-Lite & TCP & InterFuser & MTR+MPC & TF++ & CaRL & PlanT 2.0 & Diffusion-Planner\\
\midrule
42Routes & 86.4 & 20.2 & 26.2 & 28.5 & 66.4 & 72.4 & 89.6 & 29.4\\
Town05 Long & 50.7 & 14.6 & 17.4 & 9.3 & 55.3 & 73.5 & 56.1 & 14.2\\
Town05 Short & 86.9 & 45.9 & 52.0 & 31.9 & 90.2 & 83.8 & 84.1 & 51.3 \\
Longest6 & 64.0 & 7.1 & 5.9 & 12.1 & 2.5 & 67.9 & 70.6 & 15.6\\
\midrule
ParkingExit & 100.0 & 0.9 & 0.5 & 10.6 & 84.7 & 51.9 & 100.0 & 0.8\\
ControlLoss & 62.5 & 13.8 & 58.5 & 37.7 & 89.9 & 82.1 & 80.2 & 60.0\\
\midrule
SignalizedJunctionLeftTurn & 92.4 & 6.3 & 7.8 & 41.6 & 92.9 & 18.0 & 87.9 & 90.4 \\
NonSignalizedJunctionRightTurn & 70.4 & 51.6 & 57.0 & 78.8 & 100.0 & 58.5 & 100.0 & 51.3\\
NonSignalizedJunctionLeftTurn & 62.1 & 20.2 & 68.3 & 51.3 & 89.1 & 88.6 & 81.5 & 25.9\\
OppositeVehicleTakingPriority & 86.7 & 29.1 & 100.0 & 40.0 & 100.0 & 60.0 & 100.0 & 34.6\\
OppositeVehicleRunningRedLight & 33.0 & 14.6 & 5.0 & 8.7 & 14.6 & 14.6 & 5.9 & 14.6\\
\textbf{Average} & 68.9 & 24.4 & 47.6 & 44.1 & 79.3 & 47.9 & 75.1 & 43.4\\
\midrule
VehicleOpensDoorTwoWays & 27.2 & 0.0 & 2.5 & 5.9 & 13.3 & 37.9 & 18.6 & 8.9\\
ParkedObstacleTwoWays & 66.7 & 2.6 & 0.4 & 2.6 & 62.6 & 41.2 & 97.9 & 17.5\\
InvadingTurn & 64.1 & 1.0 & 1.2 & 22.0 & 67.3 & 53.4 & 71.5 & 2.0\\
ConstructionObstacle & 41.1 & 1.1 & 3.7 & 33.1 & 12.4 & 34.2 & 6.7 & 9.0\\
HazardAtSideLane & 8.2 & 8.2 & 9.2 & 23.9 & 25.8 & 46.5 & 12.5 & 92.0\\
Accident & 26.0 & 30.2 & 51.4 & 45.7 & 27.3 & 28.1 & 13.2 & 30.3 \\
ConstructionObstacleTwoWays & 18.1 & 0.8 & 0.4 & 1.6 & 27.2 & 7.3 & 8.9 & 6.8\\
\textbf{Average} & 35.9 & 6.3 & 9.8 & 19.3 & 33.7 & 35.5 & 32.8 & 23.8\\
\midrule
DynamicObjectCrossing & 57.3 & 35.2 & 41.1 & 62.7 & 60.3 & 62.7 & 43.9 & 6.8\\
VehicleTurningRoute & 60.1 & 17.0 & 20.5 & 32.0 & 60.4 & 16.3 & 63.2 & 36.0\\
VehicleTurningRoutePedestrian & 7.4 & 17.9 & 6.7 & 18.5 & 22.3 & 27.0 & 8.6 & 19.8\\
\textbf{Average} & 41.6 & 23.4 & 22.8 & 37.7 & 47.7 & 35.3 & 38.6 & 20.9\\
\bottomrule
\end{tabular}%
%}
\end{table*}

\begin{table*}[t]
\centering
%\scriptsize
\setlength{\tabcolsep}{2.2pt}
\renewcommand{\arraystretch}{0.80}
\caption{Detailed route-completion (RC) scores for each map and traffic scenario.}
\label{tab:scenario-route-completion}
%\resizebox{\textwidth}{!}{%
\begin{tabular}{lrrrrrrrr}
\toprule
Traffic scenarios or maps & PDM-Lite & TCP & InterFuser & MTR+MPC & TF++ & CaRL & PlanT 2.0  & Diffusion-Planner\\
\midrule
42Routes & 100.0 & 51.4 & 56.4 & 100.0 & 97.5 & 82.2 & 100.0 & 63.0\\
Town05 Long & 86.7 & 47.8 & 53.1 & 78.4 & 89.0 & 83.1 & 88.9 & 33.3\\
Town05 Short & 100.0 & 70.6 & 86.6 & 99.2 & 90.8 & 93.3 & 100.0 & 71.7\\
Longest6 & 98.8 & 24.0 & 20.9 & 81.5 & 4.6 & 80.0 & 99.4 & 43.6\\
\midrule
ParkingExit & 100.0 & 1.2 & 1.4 & 100.0 & 88.5 & 100.0 & 100.0 & 1.5\\
ControlLoss & 100.0 & 56.3 & 100.0 & 100.0 & 100.0 & 100.0 & 100.0 & 100\\
\midrule
SignalizedJunctionLeftTurn & 100.0 & 29.2 & 25.0 & 100.0 & 100.0 & 18.0 & 100.0 & 100.0\\
NonSignalizedJunctionRightTurn & 100.0 & 100.0 & 100.0 & 100.0 & 100.0 & 100.0 & 100.0 & 100.0\\
NonSignalizedJunctionLeftTurn & 100.0 & 72.8 & 100.0 & 100.0 & 100.0 & 100.0 & 100.0 & 32.4\\
OppositeVehicleTakingPriority & 100.0 & 74.5 & 100.0 & 100.0 & 100.0 & 100.0 & 100.0 & 67.5\\
OppositeVehicleRunningRedLight & 42.9 & 14.6 & 13.5 & 14.8 & 14.6 & 14.6 & 15.0 & 14.6\\
\textbf{Average} & 88.6 & 58.2 & 67.7 & 83.0 & 82.9 & 66.5 & 83.0 & 63.0\\
\midrule
VehicleOpensDoorTwoWays & 56.9 & 0.0 & 14.8 & 55.0 & 33.3 & 56.3 & 57.0 & 25.9\\
ParkedObstacleTwoWays & 100.0 & 4.9 & 1.6 & 16.2 & 100.0 & 100.0 & 100.0 & 68.1\\
InvadingTurn & 100.0 & 5.4 & 20.1 & 37.3 & 100.0 & 100.0 & 100.0 & 35.8\\
ConstructionObstacle & 71.7 & 3.3 & 12.6 & 100.0 & 23.0 & 66.1 & 26.0 & 67.1\\
HazardAtSideLane & 15.7 & 43.0 & 30.2 & 100.0 & 40.9 & 100.0 & 99.5 & 99.1\\
Accident & 48.1 & 53.4 & 53.3 & 45.7 & 49.0 & 53.4 & 50.0 & 51.5\\
ConstructionObstacleTwoWays & 54.9 & 4.3 & 1.0 & 6.3 & 55.7 & 34.9 & 54.8 & 20.9\\
\textbf{Average} & 63.9 & 16.3 & 19.1 & 51.5 & 57.4 & 73.0 & 69.6 & 52.6\\
\midrule
DynamicObjectCrossing & 60.3 & 59.8 & 61.2 & 62.7 & 60.4 & 62.7 & 62.7 & 20.9\\
VehicleTurningRoute & 100.0 & 37.4 & 64.8 & 100.0 & 100.0 & 16.3 & 100.0 & 61.2\\
VehicleTurningRoutePedestrian & 17.1 & 27.0 & 17.9 & 27.0 & 27.0 & 27.0 & 21.6 & 67.1\\
\textbf{Average} & 59.1 & 41.4 & 48.0 & 63.2 & 62.5 & 35.3 & 61.4 & 53.9\\
\bottomrule
\end{tabular}%
%}
\end{table*}

\begin{table*}[t]
\centering
%\scriptsize
\setlength{\tabcolsep}{2.2pt}
\renewcommand{\arraystretch}{0.80}
\caption{Detailed infraction-penalty (IP) scores for each map and traffic scenario.}
\label{tab:scenario-infraction-penalty}
%\resizebox{\textwidth}{!}{%
\begin{tabular}{lrrrrrrrr}
\toprule
Traffic scenarios or maps & PDM-Lite & TCP & InterFuser & MTR+MPC & TF++ & CaRL & PlanT 2.0  & Diffusion-Planner\\
\midrule
42Routes & 0.90 & 0.37 & 0.41 & 0.28 & 0.67 & 0.87 & 0.90 & 0.49\\
Town05 Long & 0.60 & 0.36 & 0.41 & 0.11 & 0.59 & 0.90 & 0.61 & 0.52\\
Town05 Short & 0.88 & 0.66 & 0.58 & 0.31 & 0.97 & 0.90 & 0.84 & 0.72\\
Longest6 & 0.64 & 0.38 & 0.41 & 0.16 & 0.54 & 0.83 & 0.71 & 0.47\\
\midrule
ParkingExit & 1.00 & 0.71 & 0.38 & 0.11 & 0.92 & 0.52 & 1.00 & 0.50\\
ControlLoss & 0.63 & 0.24 & 0.57 & 0.38 & 0.90 & 0.82 & 0.80 & 0.60\\
\midrule
SignalizedJunctionLeftTurn & 0.92 & 0.22 & 0.33 & 0.42 & 0.93 & 1.00 & 0.88 & 0.90\\
NonSignalizedJunctionRightTurn & 0.70 & 0.52 & 0.57 & 0.79 & 1.00 & 0.59 & 1.00 & 0.51\\
NonSignalizedJunctionLeftTurn & 0.62 & 0.26 & 0.67 & 0.51 & 0.89 & 0.89 & 0.81 & 0.80\\
OppositeVehicleTakingPriority & 0.87 & 0.39 & 1.00 & 0.40 & 1.00 & 0.60 & 1.00 & 0.51\\
OppositeVehicleRunningRedLight & 0.90 & 1.00 & 0.40 & 0.59 & 1.00 & 1.00 & 0.39 & 1.00\\
\textbf{Average} & 0.80 & 0.48 & 0.59 & 0.54 & 0.96 & 0.81 & 0.82 & 0.74\\
\midrule
VehicleOpensDoorTwoWays & 0.48 & 0.71 & 0.17 & 0.11 & 0.36 & 0.67 & 0.33 & 0.41\\
ParkedObstacleTwoWays & 0.67 & 0.54 & 0.27 & 0.16 & 0.63 & 0.41 & 0.98 & 0.26\\
InvadingTurn & 0.64 & 0.18 & 0.05 & 0.59 & 0.67 & 0.53 & 0.72 & 0.05\\
ConstructionObstacle & 0.57 & 0.36 & 0.28 & 0.33 & 0.54 & 0.52 & 0.26 & 0.14\\
HazardAtSideLane & 0.52 & 0.19 & 0.28 & 0.24 & 0.62 & 0.46 & 0.13 & 0.93\\
Accident & 0.54 & 0.56 & 0.96 & 1.00 & 0.55 & 0.53 & 0.26 & 0.59\\
ConstructionObstacleTwoWays & 0.33 & 0.20 & 0.42 & 0.25 & 0.49 & 0.21 & 0.16 & 0.32\\
\textbf{Average} & 0.54 & 0.39 & 0.35 & 0.38 & 0.55 & 0.48 & 0.41 & 0.39\\
\midrule
DynamicObjectCrossing & 0.95 & 0.59 & 0.73 & 1.00 & 1.00 & 1.00 & 0.70 & 0.59\\
VehicleTurningRoute & 0.60 & 0.45 & 0.29 & 0.32 & 0.60 & 1.00 & 0.63 & 0.35\\
VehicleTurningRoutePedestrian & 0.43 & 0.66 & 0.38 & 0.68 & 0.83 & 1.00 & 0.40 & 0.50\\
\textbf{Average} & 0.66 & 0.57 & 0.47 & 0.67 & 0.81 & 1.00 & 0.58 & 0.48\\
\bottomrule
\end{tabular}%
%}
\end{table*}

\subsection{Comparative Analysis}
\noindent \textbf{Input Representation and Evaluation Horizon.} The results indicate that privileged state information is particularly beneficial on long route-level evaluations, but does not guarantee the strongest performance in every localized scenario. At the map level, PlanT 2.0, CaRL, and PDM-Lite obtain driving scores of 75.1, 74.4, and 72.0, respectively, compared with 53.6 for the strongest sensor-based method, TF++. In contrast, TF++ is highly competitive in localized traffic scenarios, obtaining the highest category-level driving score in control recovery, traffic negotiation, and braking/lane-changing scenarios. These results suggest that privileged representations provide their clearest benefit during sustained closed-loop execution, where perception and control errors can accumulate over long rollouts, while effective multimodal sensor fusion can remain competitive for shorter and more localized interactions. 

\noindent \textbf{Sensor Modality Versus Representation and Architecture.} The sensor-based results also show that input modality alone does not explain closed-loop performance. TF++ and InterFuser both use camera and LiDAR information, yet TF++ achieves a substantially higher aggregate scenario-level driving score. TCP, which relies primarily on a front-facing camera together with vehicle measurements, performs lower still. This comparison suggests that additional sensing is not sufficient by itself: the way multimodal features are fused, the intermediate quantities represented by the network, and the interface between the learned planner and the controller materially affect closed-loop robustness.

\noindent \textbf{Planning Representation and Behavioral Compliance.}
MTR+MPC also exhibits strong route-completion behavior, especially because
its MPC objective explicitly encourages close adherence to waypoints. However,
its low infraction-penalty score indicates that waypoint tracking alone is
insufficient for robust closed-loop driving. Because the MPC formulation does not
explicitly encode all traffic-rule constraints, the controller frequently violates
traffic rules, leading to weaker driving scores despite relatively high route
completion.

Diffusion-Planner illustrates the opposite side of the progress-compliance trade-off. MTR+MPC and Diffusion-Planner obtain similar aggregate scenario-level driving scores, 30.4 and 29.8, respectively, but Diffusion-Planner has lower route completion and a higher infraction-penalty score. Thus, similar final driving scores can arise from qualitatively different failure mechanisms. This shows why aggregate leaderboard scores alone can obscure the behavioral mechanism responsible for failure.

\noindent \textbf{Training Paradigm Among Privileged Planners.} The privileged planners provide a useful comparison among rule-based planning, reinforcement learning, and imitation learning. At the map level, PlanT 2.0 achieves the highest driving score and route completion, with DS = 75.1 and RC = 97.1, whereas CaRL achieves the highest infraction penalty, IP = 0.87. PDM-Lite remains competitive with DS = 72.0 and RC = 96.4. These results suggest different progress-compliance trade-offs across training paradigms. PlanT 2.0's object-centric imitation-learning formulation and separate lateral and longitudinal planning outputs are associated with strong sustained route progress, whereas CaRL's PPO produces the strongest aggregate infraction performance. The rule-based PDM-Lite planner remains a strong privileged reference but is not universally dominant. Since these methods also differ in representation and controller design, we interpret these observations as comparative evidence rather than isolated causal effects of the training algorithm.

\noindent \textbf{Persistent Difficulty in Obstacle-Rich Scenarios.} A notable result of the scenario-level decomposition is that obstacle avoidance remains challenging across nearly all evaluated baselines. The absence of a clearly dominant architectural or training family in this category indicates that privileged state, multimodal sensing, reinforcement learning, imitation learning, rule-based planning, and trajectory generation all continue to struggle with fine-grained local interaction around unusual obstacles. This suggests that obstacle-aware local decision-making remains a shared bottleneck.

The contribution of our unified evaluation is not to employ controlled ablations, but to expose which combinations of design choices retain or lose their strengths under a common execution environment and to identify recurring behavioral trade-offs that are hidden by aggregate leaderboard rankings.

\subsection{Key Takeaways}

\begin{itemize}

 \item Privileged information primarily benefits sustained route execution. Privileged-state planners dominate the long map-level evaluations, whereas the strongest multimodal sensor-based planner remains highly competitive in several localized traffic-scenario categories.

 \item Planner output representation strongly affects the progress-compliance trade-off. High route completion does not necessarily imply robust driving: MTR+MPC demonstrates that waypoint tracking can remain strong while traffic-rule compliance is poor.

 \item Training paradigms emphasize different aspects of behavior. Among privileged planners, PlanT 2.0 provides the strongest map-level progress and overall driving score, while CaRL achieves the strongest aggregate infraction performance and PDM-Lite remains a competitive rule-based reference.

 \item Similar driving scores can hide different failure mechanisms. MTR+MPC and Diffusion-Planner achieve similar scenario-level driving scores but differ substantially in route completion and infraction penalty, showing that aggregate rankings alone are insufficient for diagnosing planner behavior.

 \item Obstacle-rich interaction remains a shared bottleneck. No evaluated family (sensor-based, privileged, rule-based, imitation-learning, reinforcement-learning, MPC-based, or diffusion-based) consistently solves the obstacle-avoidance scenarios.

\end{itemize}

\section{Conclusions}

Level 5 Autonomy promises not only personal ease of transportation, economic productivity gains on societal scale but also safety by far surpassing what we have today. 
The experiments reveal distinct trade-offs associated with information access, planning representation, training paradigm, and execution architecture. While we acknowledge that CARLA results may not perfectly generalize to the real world, we believe closed-loop simulation is a necessary and standardized first step and CARLA Leaderboard v2.1 provides the right, scalable and open source environment for rigorous evaluation. Thus, we evaluated a set of eight representative motion planning methods,
including sensor-based end-to-end models, rule-based privileged planners,
reinforcement-learning planners, object-centric transformer planners, and our
MTR+MPC hybrid. 
%We hope this paper will incentivize the AI+Autonomy research community to adopt CARLA as a training and evaluation benchmark, and make their code compatible with it. 
Our work sheds light on the gaps in these approaches, and we hope will guide the development of more robust and efficient motion planning algorithms for autonomous vehicles.

Alas, as far as self-driving cars is concerned, it seems that \textbf{\textit{we are still not there yet!}}

\bibliographystyle{IEEEtran}
\bibliography{IEEEabrv,references}

\end{document}